# Machine learning identification of organic compounds using visible light

Thulasi Bikku,[1, 2] Rubén A. Fritz,[1] Yamil J. Colón,[3] Felipe Herrera[1, 4*]

[1]Department of Physics, Universidad de Santiago de Chile, Av. Victor Jara 3493, Santiago, Chile.
[2]Computer Science and Engineering, Vignan's Nirula Institute of Technology and Science for Women, Guntur, Andhra Pradesh, India.
[3]Department of Chemical and Biomolecular Engineering, University of Notre Dame, IN, USA.
[4]ANID - Millennium Science Initiative Program, Millennium Institute for Research in Optics, Concepción, Chile.

**Abstract:**
Identifying chemical compounds is essential in several areas of science and engineering. Laser-based techniques are promising for autonomous compound detection because the optical response of materials encodes enough electronic and vibrational information for remote chemical identification. This has been exploited using the fingerprint region of infrared absorption spectra, which involves a dense set of absorption peaks that are unique to individual molecules, thus facilitating chemical identification. However, optical identification using visible light has not been realized. Using decades of experimental refractive index data in the scientific literature of pure organic compounds and polymers over a broad range of frequencies from the ultraviolet to the far-infrared, we develop a machine learning classifier that can accurately identify organic species based on a single-wavelength dispersive measurement in the visible spectral region, away from absorption resonances. The optical classifier proposed here could be applied to autonomous material identification protocols or applications.

[*]Corresponding author: felipe.herrera.u@usach.cl

# I. INTRODUCTION

Scientific data analysis has been accelerated by machine learning. Data interpretation, identification, and control of experiments by automated instruments are possible applications in this area. [1,2] Autonomous identification of chemical compounds can impact the development of compact, portable, and highly accurate sensors with lower equipment costs [3]. Machine learning models can be trained with experimental databases to classify incoming signals and predict output signals that improve the capabilities of sensing platforms [3]. Based on this output, a data-driven redesign strategy can be used to replace irrelevant system features to improve the sensor performance iteratively. This way, machine learning is an enabling tool in hardware design for intelligent sensor systems.

Remote molecular sensing based on light exploits the dispersive and absorptive response of a material system to electromagnetic radiation [4,5]. Although chemical methods can be very specific [6,7], optical sensing techniques can be more beneficial in some applications because light-matter interaction is non-destructive and can be processed remotely. The optical properties of materials such as the refractive index are fundamentally related to microscopic physicochemical properties such as the dynamic polarizability, as well as macroscopic variables such as concentration, temperature, and pressure [8,9]. The refractive index is commonly used for quantifying the content of target molecules in agricultural [10,11], pharmaceutical [8], and manufacturing applications. Therefore, a refractive index database over a broad range of frequencies can serve as useful training data for a machine learning workflow that enables the autonomous identification of chemical compounds using light.

Molecular spectroscopic databases have already been used for training machine learning algorithms in chemical identification problems [12–14]. Efforts have focused on training classifiers with infrared (IR) absorption and Raman scattering databases, in the mid-infrared (mid-IR) spectral range ($\lambda \sim 3 - 50\,\mu m$) [11–31]. In general, the information in the mid-IR is so rich and complex that it would be very unlikely for different molecules to have the same peak structure, in particular in the so-called "fingerprint" region [32,33,34]. Compounds with the same chemical formula but different spatial conformation (isomers) can thus be discriminated by analyzing the position of their Raman peaks, for example [35]. Machine learning classifiers trained on vibrational spectroscopy data can therefore be very accurate, with identification errors of a few percent or less [12,15,27–31].

Alternative machine learning strategies for molecular classification that are not trained with infrared absorption or scattering spectra have also been reported [27–31]. Away from absorption resonances, molecules and materials experience a dispersive response in the presence of external electromagnetic fields, which is determined by the real part of the dielectric function of the medium $\epsilon(\lambda)$. The imaginary part of $\epsilon(\lambda)$ gives the extinction coefficient $k(\lambda)$[1], which quantifies electromagnetic energy loss due to absorption and scattering. Absorption and dispersion are fundamentally related to each other via the Kramer-Kronig relation [9]. In frequency regions with negligible extinction, the dispersive response is entirely determined by the refractive index $n(\lambda) = \sqrt{\text{Re}[\epsilon]}$, which provides information on the electronic structure of materials ($\lambda \sim 400 - 800$ nm) or their dielectric properties ($\lambda > 50$ μm). As mentioned above, the refractive index $n(\omega)$ is also an extensive property and thus correlates with the molecular density.

The refractive index has been used for detecting tissue damage in biomedical samples using terahertz transmission images ($\lambda > 100$ μm) and deep learning, reaching recall performances of up to 93% with feature engineering techniques [14,36]. Databases with the static refractive indices ($\lambda \to \infty$) of small organic compounds, solvents and polymers have also been developed for training regression models that predict the refractive index given a set of structural and quantum chemical descriptors [37–42]. Predictive models can be useful for the automated discovery of organic optoelectronic materials [43]. Experimental databases with organic refractive indices at the sodium wavelength (589 nm) have also been used for constructing quantitative structure-property relationships (QSPR) that can be used, for example, in the chemical analysis of organic mixtures using visible light [44].

For molecular classification tasks based on IR spectral data, the target feature of interest for a classifier ("molecule identity") is learned from a high-dimensional feature vector that encodes the spectral peaks of interest [12]. In contrast, the static refractive index of organic compounds is a single-valued feature that does not have the same information content [45]. The same applies to refractive index databases at a single wavelength away from ultraviolet and infrared absorption resonances. New databases and analysis techniques are thus needed for applications that can benefit from accurate organic molecule classification protocols based on the dispersive response of materials over a tunable range of wavelengths in the visible.

Motivated by the growing interest in developing autonomous tools for organic materials discovery [46,47], here we demonstrate a *proof-of-concept* machine learning classifier of organic compounds based on the measurement of the refractive index $n(\lambda_0)$ at a single optical wavelength $\lambda_0$ anywhere in the visible spectral region ($400 - 750$ nm), where most organic compounds are fully transparent [48]. The classifier is trained with a publicly available materials science database containing the optical constants of 61 organic molecules and polymers over the spectral range spanning from ultraviolet to far-infrared wavelengths. The classification scheme is illustrated in Fig. 1. We envision the proposed classifier for data analysis at the output of a liquid or gas-phase chromatographer that can separate complex chemical mixtures into multiple single-component fractions [49].

---

[1] We follow photonics notation and denote the extinction coefficient by $k$.

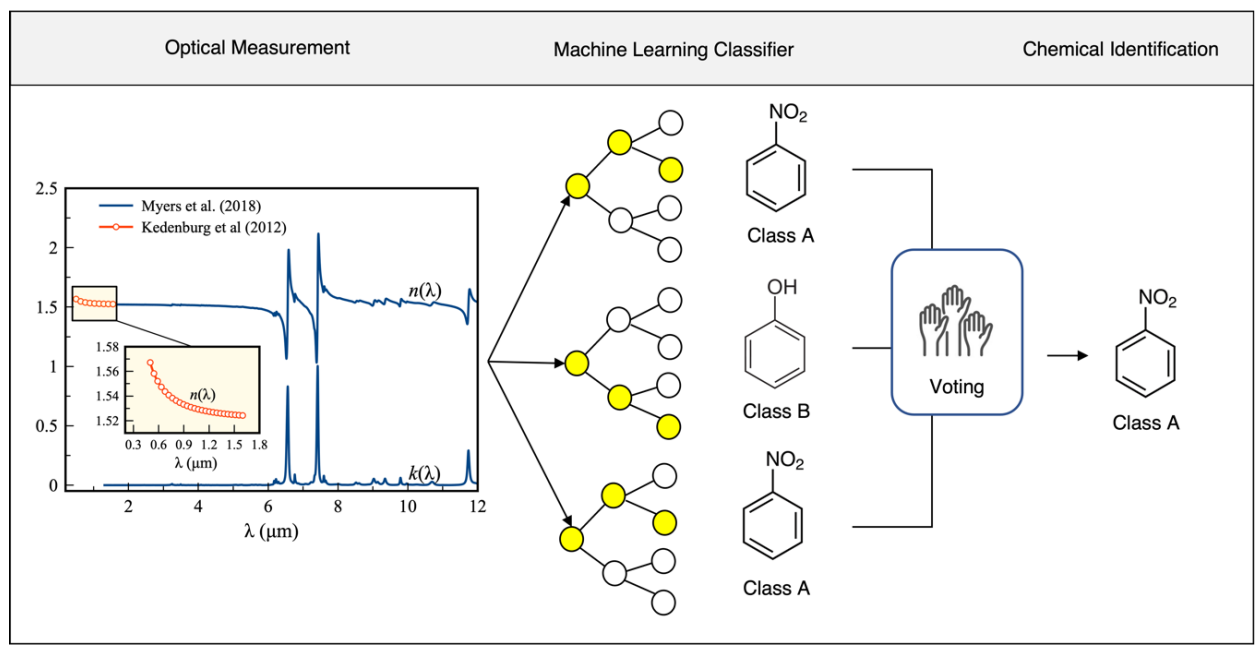

**Figure 1**. Illustration of the proposed machine learning classification scheme for chemical identification. The experimental refractive index and extinction training data in the left side are reproduced from Ref. [50].

The rest of the article is organized as follows: After describing the structure and information content of the original spectral database in Sec. II.1, we describe the data preprocessing strategies in Sec. II.2 and demonstrate classification errors smaller than 1% in the visible range in Sec. III. Comparisons with recent Raman-based classifiers are given in Sec. III.4. We conclude and suggest future directions and applications in Sec. IV.

## II. METHODS

### II.1. Experimental refractive index database

We run a web scrapping script on the publicly available database of experimental optical constants from the https://refractiveindex.info [50]. The website is a repository of published data from the scientific literature since 1940. The site is organized into categories that resemble a virtual bookshelf: The "Shelf" category groups materials into inorganic, organic, glasses, others, and 3D; the "Book" category contains the chemical compound name; the "Page" subsection refers to the source where the optical data was first reported in the literature, as well as comments and other information such as the group velocity and group velocity dispersion, the measurement wavelength range and the state of matter of the sample (gas, liquid, or solid). Each "page" record has a .csv file with the spectral data for $n$ (refractive index) and $k$ (extinction coefficient) over a range of wavelengths ($\lambda$). From the general dataset, we build a smaller set by selecting "Organic Materials" in the "Shelf" category, which contains 61 organic molecules and polymers. The compiled file has 194,816 data records, sorted in columns with the features "Shelf", "Book", "Page", "λ" (wavelength), "$n$" (refractive index), and "$k$" (extinction coefficient). The data has 418 missing $n$ values and 60,944 missing $k$ values. Example data records are shown for the "Acetone" class in Table S1 of the Supplementary Material (SM).

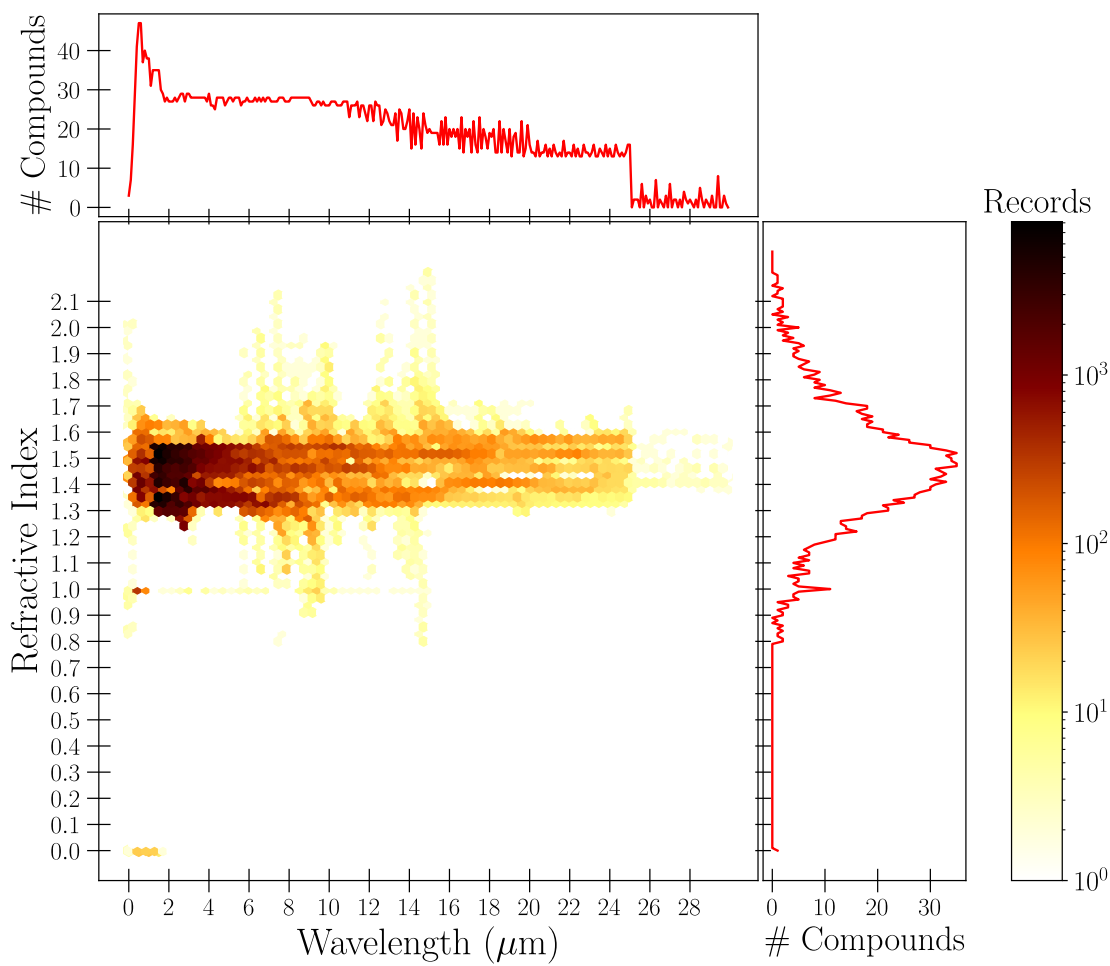

**Figure 2**: Distribution of the raw data from the https://refractiveindex.info/ [50] over a broad range of wavelengths and refractive indices. Side panels show cuts of the data record distribution over wavelengths (top) and refractive indices (right).

In Fig. 2, we show a visualization of the organic dataset over a grid of wavelengths ($\lambda$) and refractive indices ($n$). The color code indicates the number of data records in each ($\lambda$, $n$) region [ The records involve measurements up to 25 μm and refractive indices in the range (0.0 − 2.3)]. The upper and right-side panels show the number of organic compounds (counted as a label on the "Book" category) with data records in each wavelength and refractive index range.

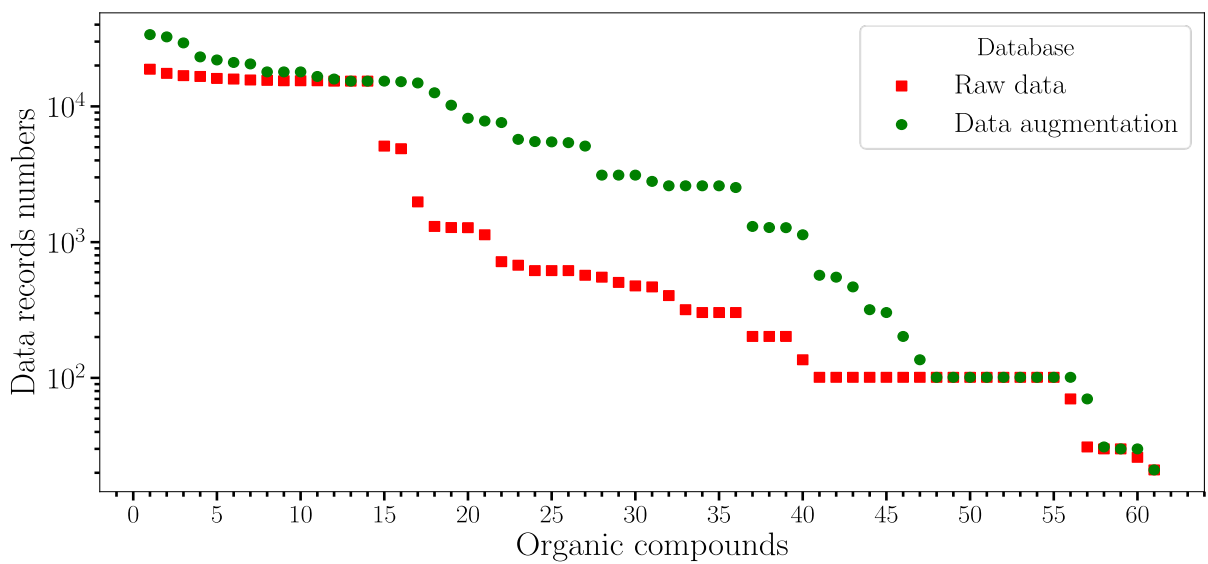

**Figure 3**: Number of data records per molecular class (61 classes) available in the raw database from Ref. [50]. The record distribution after data augmentation via Sellmeier fitting in the UV/VIS is also shown. The correspondence between the class labels and organic compound names is given Table S5 of the Supplementary Material.

## II.2. Preprocessing the training dataset

The original database is naturally heterogeneous and imbalanced, as some compounds and frequency regions have been studied more intensely than others in the literature. This represents a challenge for the implementation of machine learning classification algorithms [51]. Datasets with unequal data records per target feature lead to majority and minority classes, thus affecting the overall predictive accuracy of classification models [52]. Figure 3 shows the number of data records per organic molecule class for the raw database (RD) and after pre-processing through data augmentation (DA). The number of records in the original data vary from only few spectral points (e.g., Polymethyl Pentene, Class Label = 58) to a few thousands (e.g, Ethanol, Class Label = 1), which illustrates the class imbalance problem.

We tried several pre-processing strategies to overcome class imbalance in the raw dataset, including oversampling (OS), undersampling (US) and physics-based DA. We also used feature engineering (FE-1 and FE-2) strategies, spectral-based binning (SBB), and spectral-based binning with feature engineering (SBB-FE1 and SBB-FE2), attempting to increase the prediction accuracies with the original imbalanced dataset. The details of these pre-processing strategies are given in the Results section.

## II.3. Random forest classifier

The random forest (RF) algorithm was chosen as the default method for classification. Random forest is a supervised machine learning algorithm based on an ensemble of decisions trees [53]. It uses bagging and feature randomness when building each individual decision tree to create an uncorrelated forest of trees whose prediction is more accurate by committee than individually. The dataset was divided into 75% for training and 25% for testing. The results below were obtained with a single instance of the split, but additional tests on the raw database show that the accuracies do not vary significantly when averaging over different splits (see Table S2 in the SM). RF was implemented using Python's Scikit-Learn library with default hyperparameters [54]. In early stages of this study, we obtained classification accuracies with alternative models such as Gradient Boosting, Support Vector Classification, and Logistic Regression on the raw dataset, and random forest performed best (see test results Table S4 in the SM). The code and the datasets used in this work are publicly available at  https://github.com/fherreralab/organic_optical_classifier.

## III. RESULTS AND DISCUSSION

In this Section we discuss the model performance obtained for classification trained with the original imbalanced dataset and after preprocessing the data. We tested several data preprocessing techniques to gain insight on the optimal experimental setup necessary for achieving reasonable identification outcomes using minimal optical measurements.

## III.1. Classification accuracies with imbalanced class sets

Incorporating domain knowledge into the features of the training set is known to increase the prediction performance of classification models. Based on this intuition and inspired by the feature structure of Raman-based classifiers [17,23,25] where the target class is associated with a high-dimensional feature vector containing the spectral peaks, we tested whether grouping small fractions of the curves $n(\lambda)$ and $k(\lambda)$ into feature vectors $\boldsymbol{x} = [\lambda_j, n_j, k_j]$ improved the classifier performance. As explained above, the original dataset has a three-dimensional feature vector per target class ($j = 1$) representing a single evaluation of the $n$ and $k$ curves at a given value of $\lambda$. We then build six-dimensional ($j = 1,2$) and nine-dimensional feature vectors ($j = 1,2,3$) containing the information from two and three consecutive points in the $n$ and $k$ curves. We refer to the six- and nine-dimensional feature vector schemes as Feature Engineering 1 (FE1) and 2 (FE2), respectively. As a result of the increase in the number of features per target class, the number of records is reduced.  For classification problems with large datasets, binning strategies can prove useful for improving the overall class prediction accuracy [55,56]. In addition to the FE1 and FE2 strategies, we adopt a spectral-based binning (SBB) strategy based on the division of the wavelength domain into five spectroscopic regions: UV [$\lambda < 0.40\ \mu$m] containing 1,773 data records, Visible (VIS) [$0.40\,\mu m < \lambda < 0.75\,\mu m$] with 5,979 records; Near Infrared (Near-IR) [$0.75\,\mu m < \lambda < 1.50\,\mu m$] with 35,445 records; Infrared (IR) [$1.50\,\mu m < \lambda < 4.0\,\mu m$] with 135,407 records; and Far-Infrared (Far-IR) [$\lambda > 4.0\,\mu m$] with 66,678 records. Our splitting of the IR spectrum into sub-regions is not necessarily standard [57].

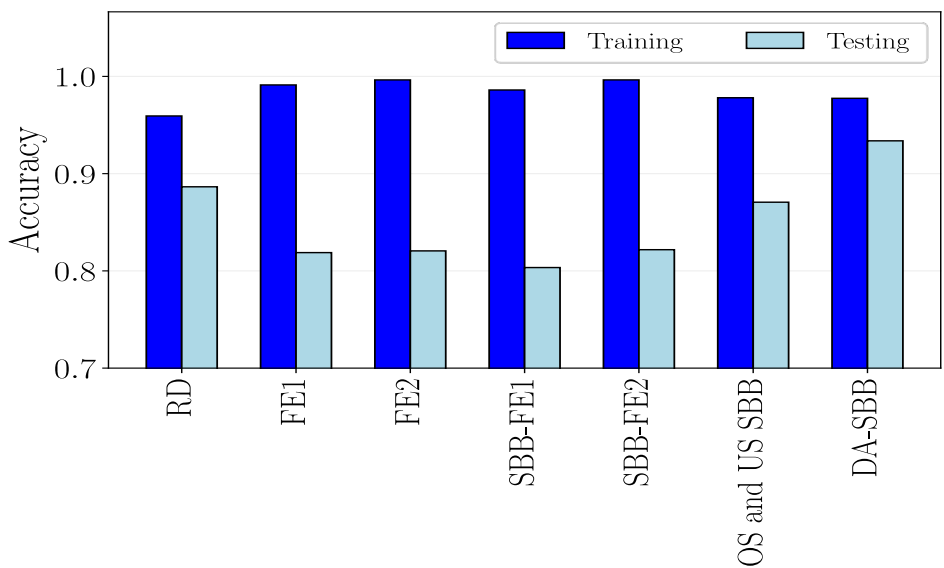

**Figure 4**: Overall training and testing accuracy for imbalanced data, not separated by wavelength range for the different data preprocessing strategies.

In Fig. 4, we compare the training and testing accuracies of the random forest classification training with the raw database (RD), as well as FE and SBB pre-processing strategies. Accuracies are also provided in Tables S2 and S3 of the SM. The overall testing accuracies are about 80% with and without preprocessing, although feature engineering and spectral-based binning tend to reduce the prediction accuracy. Combining FE and SBB did not improve performance relative to the raw data. The low performance of the classifier over the entire range of wavelengths (UV to Far-IR) comes roughly speaking from an average of spectral regions of very high accuracies (IR) and regions with very low accuracies (UV-VIS). In what follows, we separately studied the performance in different spectral regions.

### III.2. Addressing class imbalance in the database

In Figs. 3 and 5 we illustrate that the original dataset contains a disproportionate number of data records per organic compound in the IR spectral bin ($1.50 - 4.0\,\mu m$), relative to the UV and VIS bins. This type of record distribution generates a class imbalance problem [58] that we address using the following strategies: undersampling (US), oversampling (OS), and data augmentation (DA).

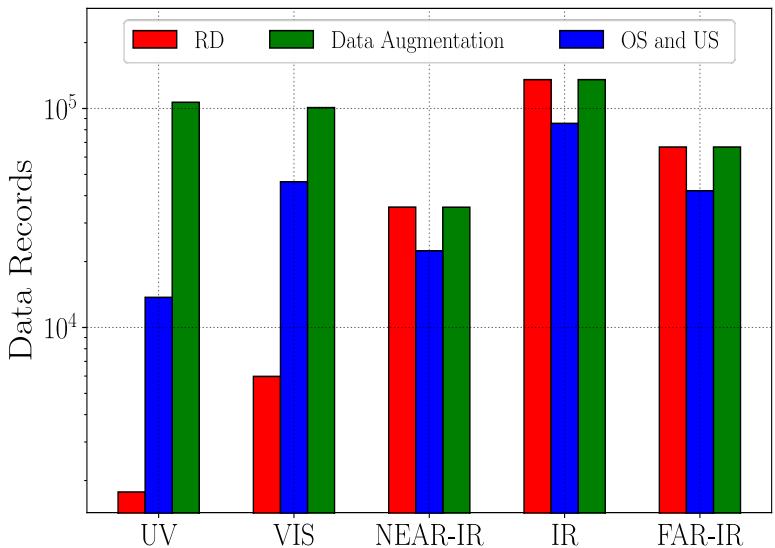

**Figure 5**: Distribution of data records per spectral bin in the raw database (RD), after Oversampling (OS), Undersampling (US), and Data augmentation (DA).

Resampling strategies aim to balance classes in the training data by reshaping the dataset such that the number of records in the different classes becomes comparable [58]. Undersampling is a method that randomly reduces the number of data records in the majority class, while oversampling duplicates randomly chosen records in the minority class. Both resampling strategies can be effective when used independently or combined. Figure 4 shows the record distribution used for training the random forest model after OS and US are carried out. In the resampled dataset, the number of records in the UV/VIS is comparable with the IR.

In addition to resampling (OS and US), we augment the dataset using physics-based modeling. Specifically, we generate refractive index data using the Sellmeier equation [9]

$$n^2(\lambda) = A + \frac{B_1 \lambda^2}{\lambda^2 - C_1} + \frac{B_2 \lambda^2}{\lambda^2 - C_2}, \qquad (1)$$

where $(A, B_j, C_j)$ are phenomenological coefficients. For each molecule in the original database that has refractive index measurements in the UV-visible, we fit the experimental curves for $n^2$ using Eq. (1) to obtain Sellmeier coefficients. These are then used to interpolate the index data in the region $\lambda = [200,750]$ nm. Therefore, we exclude the near-infrared wavelengths from the fitting, despite the Sellmeier equation still being valid in this region for most organic materials. This data augmentation procedure increases the number of records in the UV and VIS bins to about 3,000 additional points per chemical compound. Figure 5 shows that the distribution of training records set after DA changed significantly with respect to RD. The augmented dataset has about 400,000 records, with comparable numbers of records in all spectral bins. DA is used in ML to resolve the imbalance problem, and it has been used in Raman-based classification problem in Refs. [27–31].

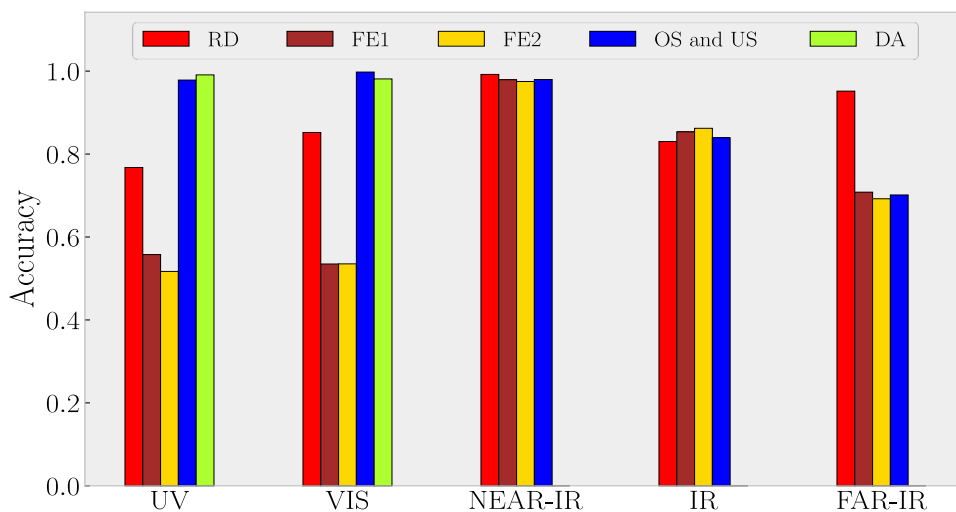

**Figure 6**: Testing accuracy after the spectral-based binning with different data preprocessing strategies: Featured engineering (FE1 and FE2), oversampling (OS), undersampling (US), and data augmentation (DA). The accuracies of the raw database are also shown for comparison.

After addressing class imbalance in the raw dataset, we tested the prediction performance of the random forest classifier on the different spectral bins. Figure 6 shows that the accuracies after US and OS improved significantly relative to the imbalanced dataset, reaching 97% for UV and 99% for VIS regions (see also Table S3 in the SM). The accuracies after US and OS however did not change significantly in the Near-IR (2% improvement) and IR (0% improvement) bins. There is a decrease of 25% in accuracy for Far-IR, likely due to the reduction of data records on this region during undersampling. On the other hand, data augmentation (DA) significantly improved the performance of the classifier (99% for UV and 98% for VIS) without affecting the accuracies in the infrared bins (see Table S3 in the SM). The accuracies for FE1 and FE2 significantly decreased as expected. The reduced number of data records resulting from the grouping of the features dramatically affect the model performance in the less represented spectral bins.

The results in Fig. 6 suggest that measuring the refractive index in the UV-VIS region is in principle sufficient for a precise classification of chemical compounds. In comparison with the accuracies reached in the far infrared spectrum ($\lambda > 5$ μm, fingerprint region), the classification accuracies in the visible region after data augmentation are equally good, without additional steps of hyperparameter optimization for the random forest model.

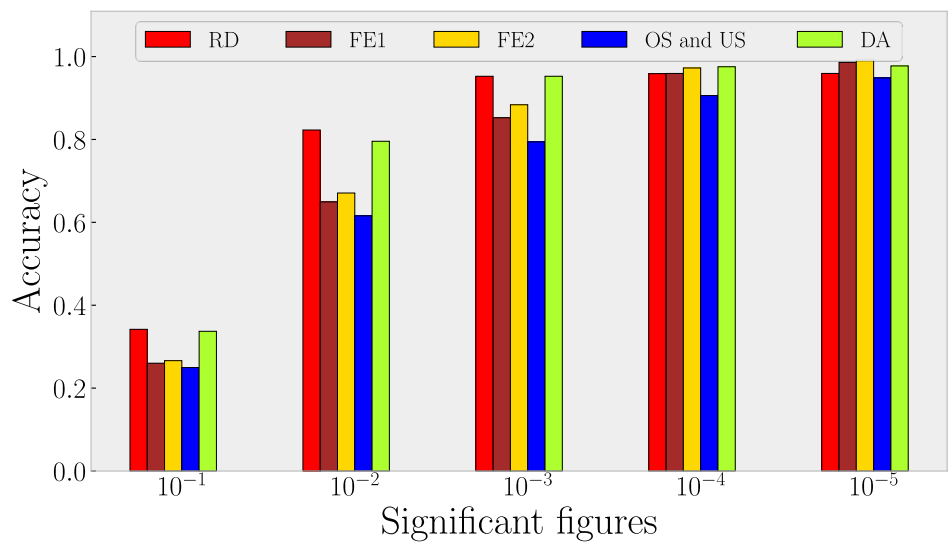

**Figure 7.**- Testing accuracies for data truncated to a fixed number of significant figures. Results are shown for different preprocessing strategies: Featured Engineering (FE1 and FE2), oversampling (OS), undersampling (US), and data augmentation (DA). The accuracies for the raw database (RD) are also shown for comparison

### III.3. Estimating the required measurement precision

The precision with which the refractive index is measured in experiments should affect the classifier performance. Compounds with similar refractive indices at a given wavelength in the visible would not be reliably distinguished if the index measurement does not have enough significant digits. We quantify this intuition by computing the classification accuracy achieved by the random forest model using training data truncated to a finite number of significant figures. In Fig. 7 we show the testing accuracies obtained using different decimal digits for $n(\lambda)$ in the training set, separating the results by preprocessing strategy. While the classification accuracies are in general poor (<82%) for index measurements with two significant figures or less, the raw database already gives better classification accuracies (~95%) when trained with index data of at least three decimal places. Increasing the number of significant figures beyond four decimals does not significantly improve the accuracy, regardless of the preprocessing strategy used.

### III.4.- Comparison with Raman-based classifiers

Class imbalance has also been addressed in other molecular classification studies based on Raman spectral databases, which also tend to be heterogeneous with a data record distribution that overrepresents compounds of particular interest in chemistry [27–31]. To address class imbalance, researchers have explored data-augmentation strategies such as peak shifting, noise addition, smoothing, spline interpolation, and polynomial reconstruction. These pre-processing strategies were implemented to reconstruct the dataset before training deep learning classifier models. In our analysis of refractiveindex.info database, the Sellmeier fitting procedure is a valid augmentation strategy that can be used to reconstruct the part of the training set corresponding to ultraviolet and visible wavelengths, without introducing data leakage [9]. In Table 1 we compare the refractive index approach with the recent Raman-based classifiers in the literature, showing that the dispersive method can also give high classification performance, using a similar a pre-processing strategy (interpolation) on datasets with a comparable volume of data records and molecular classes [27–31].

| Method | Spectral Database | Testing Accuracies (%) | Data Augmentation | Ref. |
|---|---|---|---|---|
| 1D-CNN | Minerals and organic compounds. | 100 | Standard augmentation transformations in vibrational spectroscopy. | [30] |
| CRL | 72 organic compounds | 97.5 | Gaussian noise and linear combination. | [27] |
| DNN | 72 organic compounds | 92.6* 96.4** | Shifting, Gaussian noise, and interpolation | [29] |
| CNN | 72 organic compounds | 81.9* 86.0** | Shifting, Gaussian noise, and interpolation. | [29] |
| DRCNN | 72 organic compounds | 98.1 | Shifting and Gaussian noise. | [28] |
| 1D-CNN and KNN | 620 mineral and 211 synthetic organic pigments | 97.38 | Shifting, Gaussian noise, Savitzky-Golay smoothing, spline interpolation, and polynomial reconstruction | [31] |
| RF | 61 organic compounds | 99 (UV) 98.1 (VIS) 99.2 (Near-IR) 83.1 (IR) 94.8 (Far-IR) | Sellmeier Equation fitting on UV/VIS optical ranges. | This work |

**Table 1**.- Prediction accuracies of several works implemented on database with Raman chemical spectra for comparison with this work. * No transfer learning, ** with transfer learning. The number of molecules present in the database is specified when the data is clear in the literature. We also list the data augmentation methodologies used in these works. 1D-CNN= 1D convolutional neural network, CRL= Contrastive representation learning, DNN= Deep Neural Network, CNN= Convolutional Neural Network, DRCNN = Deeply-recursive convolutional neural network, KNN= K-nearest neighbor classifier, RF= Random Forest

## IV. CONCLUSIONS

We built a machine learning classification scheme to identify organic compounds based on refractive index measurements in the visible spectral region, in which most organic compounds are highly transparent. We trained a random forest classifier using decades of experimental data from the scientific literature. The database contains 194,816 spectral records of refractive index and extinction curves of 61 organic compounds and polymers over a broad range of wavelengths from the UV to the far-infrared. There is a class imbalance problem in the experimental data that restricts the classification accuracy for refractive index inputs at visible wavelengths (400-750 nm) to approximately 80%. This imbalance is primarily due to the disproportionate number of infrared absorption records reported in the mid- and far-infrared regions. Imbalance is a common problem when working with spectroscopic databases as the experimental data is deposited from different sources [27–31].

We addressed this class imbalance issue by preprocessing the raw data before training the classifier using resampling and physics-based data augmentation strategies analogous to those employed by other ML or AI classifiers based on Raman spectra [27–31]. By training the random forest model with preprocessed balanced data, we achieve molecular classification testing accuracies in the UV and visible regions better than 98%. Additional improvements can be expected with additional steps of model hyperparameter optimization. Such high accuracies are comparable to those obtained using only Raman spectroscopy databases (see Table 1), thus demonstrating the feasibility of using machine learning tools for enhancing the capabilities of laser-based chemical sensing devices. Additional work is needed to expand and generalize the classifier to identify the structural and other chemical features of the molecules that are present in the Refractive Index Database [50]. This work serves as a starting point for the development of remote chemical sensors based on laser light.

## SUPPLEMENTARY MATERIAL

See the Supplementary Material for the list of organic compound classes in the training database, additional random forest classification tests, a visualization of the trained trees, and classification metrics using other machine learning models.

### Data availability statement
The data that support the findings of this study are openly available at
http://doi.org/10.5281/zenodo.6419970.

### Conflict of Interest
The authors have no conflicts to disclose.

**ACKNOWLEDGMENTS**

RAF is supported by DICYT-USACH grant POSTDOC USA1956_DICYT. FH and TB are supported by ANID through grants FONDECYT Regular No. 1181743 and Millennium Science Initiative Program ICN17-012. YJC thanks the University of Notre Dame for financial support through start-up funds.